%% file: NRM.tex
\documentclass[a4paper]{article}
\usepackage{iwslt15,amssymb,amsmath,epsfig}
\usepackage{algorithm}
\usepackage{algorithmic}
\usepackage{booktabs}
\usepackage{pgfplots}
\usepackage{tikz}
\usepackage{tkz-graph}

\def\layersep{2.5cm}

\def\reg{{\rm\ooalign{\hfil
     \raise.07ex\hbox{\scriptsize R}\hfil\crcr\mathhexbox20D}}}

\title{Continuous Space Reordering Models for Phrase-based MT}

\name{Nadir Durrani \hspace{15mm} Fahim Dalvi
\address{Qatar Computing Research Institute -- HBKU \\
{\small \tt \{ndurrani,faimaduddin\}@qf.org.qa
}
}
}

\usepackage{color}

\begin{document}
\maketitle
\begin{abstract}

Bilingual sequence models improve phrase-based  translation and reordering by overcoming phrasal independence assumption and handling long range reordering. However, due to data sparsity, these models often fall back to very small context sizes. This problem has been previously addressed by learning sequences over generalized representations such as POS tags or word clusters. In this paper, we explore an alternative based on neural network models. More concretely we train neuralized versions of lexicalized reordering \cite{tillman:2004:HLTNAACL} and the operation sequence models \cite{durrani-EtAl:2013:Short} using feed-forward neural network. Our results show improvements of up to 0.6 and 0.5 BLEU points on top of the baseline German$\rightarrow$English and English$\rightarrow$German systems. We also observed improvements compared to the systems that used POS tags and word clusters to train these models. Because we modify the bilingual corpus to integrate reordering operations, this allows us to also train a {\it sequence-to-sequence} neural MT model having explicit reordering triggers. Our motivation was to directly enable reordering information in the encoder-decoder framework, which otherwise relies solely on the {\tt attention model} to handle long range reordering. We tried both coarser and fine-grained reordering operations. However, these experiments did not yield any improvements over the baseline Neural MT systems. 

\end{abstract}

\section{Introduction}

Source-target bilingual sequence models 
have been used successfully as feature in phrase-based SMT \cite{zhang-EtAl:2013:NAACL-HLT,durrani-EtAl:2013:Short}. They are based on minimal translation units, and overcome independence assumption by handling non-local dependencies across phrasal boundaries, thus providing better translation and reordering mechanism. Such models however suffer from data sparsity and fall back to very small context sizes during test time. This shortcoming is addressed by learning factored models \cite{koehn-hoang:2007:EMNLP-CoNLL2007,niehues-EtAl:2011:WMT,durrani-EtAl:2013:WMT2,durrani-EtAl:2014:Coling}, learned over POS and morphological tags or using  word classes \cite{chahuneau13morphogen,wuebker-EtAl:2013:EMNLP,bisazza-monz:2014:Coling}.\footnote{as obtained during GIZA training \cite{Och:2003} or using brown clusters \cite{Brown92class-basedn-gram}} 

An alternative way to address data sparsity and learn better generalizations is to use continuous representations. Neural networks (NN) have shown success in Statistical Machine translation with n-best re-ranking \cite{schwenk:2012,le-allauzen-yvon:2012:NAACL-HLT} or directly as a feature \cite{Vaswani_2013_emnlp,Devlin_2014_acl} used during decoding. More recently, attention-based encoder-decoder Recurrant Neural Network (RNN) model \cite{bahdanau:ICLR:2015}, which trains a single large neural network, has emerged as the new state-of-the-art in MT. 

In this work, we neuralize two commonly used reordering models namely lexicalized reordering \cite{tillman:2004:HLTNAACL} and the operation sequence model (OSM) \cite{durrani-EtAl:2013:Short} and integrate them as feature in phrase-based MT. We convert word-aligned bi-text into a sequence of operations through a deterministic algorithm (See Algorithm 1 in \cite{durrani15:cl}), the resulting vocabulary and number of model parameters can become very large. A model trained on such representation may suffer from data sparsity. To overcome this, we separate the streams of source and target sequences and concatenate them to simulate the jointness. A feed-forward neural network is then trained on such concatenated n-gram sequences. 

The OSM model exhibit very rich reordering operations varying from {\tt Insert GAP} to {\tt JUMP Forward} and {\tt JUMP Backward} to multiple open gaps which may be hierarchically created. In an alternative method, we replace complicated reordering operations with {\tt Monotonic, Swap and Discontinuous} operations, and train a neural model with coarser tags. This model is similar to the lexicalized reordering model, however much richer as it is conditioned on longer source-target contextual history and also previous reordering decisions. 

We experimented with German-to-English and English-to-German language pairs. German is syntactically divergent from English and also exhibit very rich morphology, thus prone to data sparsity. These are the two problems we are addressing in this work. Our results show improvements of up to +0.6 and +0.5 BLEU points in German-to-English and English-to-German baselines respectively. We also demonstrated that neuralized OSM model performed better than the ones trained on POS tags and word-clusters. The neuralized OSM model outperformed the simpler lexicalized variant, although only slightly.

While training the Neural OSM (and Neural lexicalized reordering model) we embed reordering information in form of operations in the training corpus. This also allows us to train {\tt sequence-to-sequence} neural MT system, where the target side is conditioned on both lexical and reordering states.  \cite{cohn-EtAl:2016:N16-1} recently showed that integrating structural bias such as {\tt Position bias}, i.e. relative positions of a source and corresponding target word, improves the attention mechanism. We tried to replicate this effect by i) linearizing the source to be in the same order as the target using word-alignments and ii) incorporating reordering states. Our motivation was that such reordering triggers will aid the attention model to better handle reordering. However, our results did not yield any improvements. 

The remainder of this paper has been organized as follows: Section \ref{sec:RM} describes the operation sequence and the lexicalized reordering model. We then present the neuralized versions of these models. Sections \ref{sec:exp} and \ref{sec:nmt} describe our experimental setup and discusses the results. Section \ref{sec:rw} gives account on related work and Section \ref{sec:conc} concludes the paper.

\section{Reordering Models}
\label{sec:RM}

In this section we briefly revisit the two commonly used reordering models in the phrase-based Moses \cite{Moses:2007} namely the lexicalized reordering model and the operation sequence model. We then describe our neural versions of these models.

\subsection{Lexicalized Reordering Model}

\begin{figure}
\centering
\includegraphics[scale=0.55]{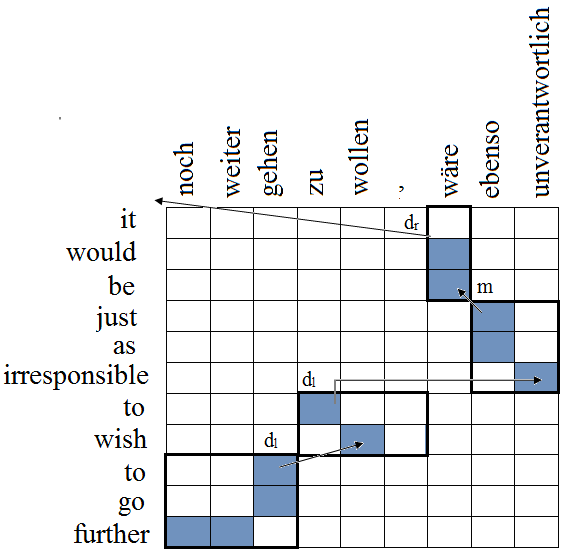}
\caption{Lexical Reordering Models \cite{IWSLT:2005}: m = monotonic, s = swap, d = discontinuity (l: left, r: right)}
\label{fig:pb-reorder}
\end{figure}

The lexicalized reordering model originally proposed by \cite{tillman:2004:HLTNAACL} is the defacto reordering model used in phrase-based SMT (PBSMT). The idea is to learn orientation of a phrase w.r.t to previous phrase (or the last word of the previous phrase). An orientation could be one of the three reordering operations namely {\tt {\bf M}onotonic, {\bf S}wap, {\bf D}iscontinuous}. If the source phrases $F_{a(j-1)}$ and $F_{a(j)}$\footnote{The mapping function $a(j)$ aligns the target phrase $E_j$ to the source phrase $F_i$, where $F_i = F_{a(j)}$.} are adjacent and in the same order as the target phrases $E_{j-1}$ and $E_{j}$, the orientation is {\tt Monotonic}. If they are in the opposite order of $E_{j-1}$ and $E_{j}$, then the orientation is {\tt Swap}, otherwise it is {\tt Discontinuous}. See Figure \ref{fig:pb-reorder} for illustration.
For each phrase, we compute its probability of being reordered with the orientations $o={M,S,D}$ as below:

\vspace{-2mm}
$$ p_r(o|F_{a(j)},E_j) = \frac{count(o, F_{a(j)},E_j)}{count(F_{a(j)},E_j})$$

Improved versions \cite{IWSLT:2005,galley-manning:2008:EMNLP} have been subsequently integrated into Moses toolkit. The former computes orientation only based on the last word of the previous phrase, rather than the entire phrase and the latter, hierarchically combines all previous phrases to compute the probability. In our work, we will compute orientation based on previous source word, but condition on $n$ previous source-target units. This is because our model is based on minimal translation units \cite{zhang-EtAl:2013:NAACL-HLT} and does not contain phrasal boundaries.

\subsection{Operation Sequence Model}

The operation sequence model (OSM) converts aligned bilingual corpus into a sequence of operations using a deterministic algorithm. 
An operation is either joint source-target lexical generation, or a reordering operation such as {\tt Insert Gap} or {\tt Jump Forward} or {\tt Backward} to a specific open gap. A Markov model is estimated from the resulting operation corpus. More formally a bilingual sentence pair $(T,S)$ and its word-alignment $A$ is transformed deterministically to a heterogeneous sequence of translation and reordering operations ($o_1,o_2,\dots,o_J$). A 5-gram model is then learned over these sequences:

\vspace{-5mm}
$$ P_{osm}(T,S) \approx \prod_{j=1}^J P(o_j|o_{j-n+1} ... o_{j-1}) $$

The operation sequence for the example shown in Figure \ref{fig:pb-reorder} according to the algorithm described in 
the original paper is given below: 

\vspace{2mm}

\textit{Generate Target Only (it) -- Insert Gap -- Generate (w\"are, would be) -- Generate (ebenso, just as) -- Generate (unverantwortlich, irresponsible) -- Jump Back (1) -- Insert Gap -- Generate (zu, to) -- Generate (wollen, wish) -- Generate Source Only (,) --  Jump Back (1) -- Insert Gap -- Generate (gehen, to go) -- Jump Back (1) -- Generate (noch weiter, further)}
\vspace{2mm}

\noindent The OSM is trained on minimal translation units (MTUs) and does not adhere to phrasal boundaries. Access to joint source target information enables it to better handle long distance dependencies. The jumps and gap operations allow OSM to learn more complex reordering patterns. However, due to data sparsity it is impossible to observe all possible reordering patterns during the training. The model therefore falls back to very small context sizes. Earlier work has addressed this problem by estimating estimating the bilingual sequence models on POS tags or word clusters \cite{crego-yvon:2010:POSTERS,niehues-EtAl:2011:WMT,durrani-EtAl:2014:Coling}.

\subsection{Neural Reordering Models}

\input{figure2.tex}

In this paper we take a different approach to address the problem of data sparsity by training the model using a feed forward neural network. Below we present the proposed neural versions of the OSM and lexicalized reordering models.

\begin{table*}
\centering
\begin{tabular}{l|l|l|l|l}
\toprule
{\bf Operations} & {\bf Source Stream} & {\bf Target Stream} & {\bf Source Stream} & {\bf Target Stream}\\
\midrule
Generate Target Only (it) & it & & it & \\
Insert Gap & Insert Gap & Insert Gap & Jump Fwd & FD \\
Generate (w\"are, would be) & w\"are & would be & w\"are & would be\\
Generate (ebenso, just as) & ebenso & just as & ebenso & just as \\
Generate (unverantwortlich, irresponsible) & unverantwortlich & irresponsible & unverantwortlich & irresponsible \\
Jump Back (1) & Jump Back (1) & Jump Back (1) & BD & BD \\
Insert Gap & Insert Gap & Insert Gap \\
Generate (zu, to) & zu & to  & zu & to \\
Generate (wollen, wish) & wollen & wish & wollen & wish \\
Generate Source Only (,) & , &  & , & \\
Jump Back (1) & Jump Back (1) & Jump Back (1) & BD & BD \\
Insert Gap & Insert Gap & Insert Gap \\
Generate (gehen, to go) & gehen & to go & gehen & to go \\
Jump Back (1) & Jump Back (1) & Jump Back (1) & BD & BD \\
Generate (noch weiter, further) & noch weiter & further & noch weiter & further \\
\bottomrule
\end{tabular}
\caption{Operation Sequences and corresponding streams for Neural OSM and Lexicalized RM training}
\label{tab:operations} 
\end{table*}

\subsubsection{Neural Operation Sequence Model}

A straight forward way is to build a neural language model using the generated sequences of operations. However, because of the joint nature of the model, the vocab size becomes quadratic ($M \times $N) causing severe data sparsity. A way to alleviate this problem is to separate out source and target streams and concatenate them to form history. See Table \ref{tab:operations} for mapping operations into separate streams of source and target operations. Here are the considerations that we made: i) When a source or target word is unaligned ({\tt Generate Source Only (Y)} or {\tt Generate Target Only (X)} operations), we don't append anything on the other side, ii) Whenever there is a reordering operation ({\tt Insert Gap/Jump Forward/Jump Back (N)}) we append it on both sides, iii) We replace source words on both sides for the {\tt Generate Self} operation, iv) Multi-word source and target cepts are collapsed together even if they appear in a different order in the original sequence, v) Note that source-side is now reordered to be order of target just as in the original model. 
We generate separate streams of source and target operation and then concatenate them to train the neural model. Let $s_o = s_{o_1}, s_{o_2} ... s_{o_n}$ and $t_o = t_{o_1}, t_{o_2} ... t_{o_m}$ be streams of source and target operations, the model is defined as:

\vspace{-4mm}
$$ P(T,S) \approx \prod_{j=1}^J P(t_{o_j}|t_{o_{j-n+1}}...t_{o_{j-1}}, s_{o_j}...s_{o_{j-m+1}}) $$
\vspace{-2mm}

\noindent where $m$ and $n$ are the source and target word histories which we concatenate to form input to the neural network. As exemplified in Figure \ref{fig:NN}, this is essentially an $(m+n)$-gram neural network LM (NNLM) originally proposed by \cite{Bengio03}. Each input word i.e. source or target vocabulary word or a reordering operation in the context is represented by a $D$ dimensional vector in the shared look-up layer $L$ $\in$ $\mathbb{R}^{|V_i| \times D}$ where $V_i$ is the input vocabulary.\footnote{Note that $L$ is a model parameter to be learned.} The look-up layer then creates a context vector $\mathbf{x_n}$ representing the context words of the $(m+n)$-gram sequence by concatenating their respective vectors in $L$. The concatenated vector is then passed through non-linear hidden layers to learn a high-level representation, which is in turn fed to the output layer. The output layer has a \texttt{softmax} activation over the output vocabulary $V_o$ of target words. Formally, the probability of getting $k$-th word in the output given the context $\mathbf{x_n}$ can be written as:

\vspace{-1mm}
\small
\begin{equation}
P(y_n = k|\mathbf{x_n}, \theta) = \frac{exp~(\mathbf{w}_k^T\mathbf{\phi (x_n)})} {\sum_{m=1}^{|V_o|} exp~({\mathbf{w}_m^T\mathbf{\phi(x_n)})}} \label{nnjm-softmax}
\end{equation}
\normalsize

\noindent where $\mathbf{\phi(x_n)}$ defines the transformations of $\mathbf{x_n}$ through the hidden layers, and $\mathbf{w}_k$ are the weights from the last hidden layer to the output layer. For notational simplicity, henceforth we will use $(\mathbf{x_n}, y_n)$ to represent a training sequence. By setting $m$ and $n$ to be sufficiently large, neural OSM can capture long-range cross-lingual dependencies between words, while still overcoming the data sparseness issue by virtue of its  distributed representations (i.e., word vectors).

\subsubsection{Neural Lexicalized Reordering Model}

We train the neural lexicalized reordering model in the same manner as that of the Neural OSM model. Traditional lexicalized reordering models use {\tt Monotonic, Swap and Discontinuous}. We retained the {\tt Swap (SW)} operation and divided the {\tt Discontinuous (D)} category into {\tt Forward Discontinuity (FD)} and {\tt Backward Discontinuity (BD)} following \cite{nagata-EtAl:2006:COLACL2}. We also removed the {\tt Monotonic} orientation from the generation as it is obvious that words flow monotonically when there is no reordering. This is also done similarly in the OSM generation. Again like the Neural OSM generation, the reordering tags are split across both the streams. See Table \ref{tab:operations} for the sample generation (last 2 columns).

Note that this model is not exactly the neural version of the lexicalized reordering in which the task is just to predict orientation/reordering decision ({\tt Monotonic, Swap, Discontinuous}) based on previous source-target word (or phrase). Here we are trying to score the entire sequence which contains both lexical (word generation) and reordering choices. The task is to find most probable sequence of lexical and reordering decisions. The difference compared to the OSM is the granularity of the reordering tags. In this model, we just have one reordering decision per lexical generation. In the OSM model, the model can have very complex sequence of reordering operations in between adjacent lexical generations. A more accurate version of the neural lexicalized reordering is described in \cite{li-EtAl:2014:Coling3}. They cast it as a classification problem, and use a continuous space representation treating a phrase as a dense real-valued vector. But unlike traditional model, they condition reordering probabilities also on the words of previous phrase to capture longer dependencies. This is similar to our work, except that our context information can go even beyond previous phrases and previous reordering decisions are also part of the context.

\subsubsection{Neural Lexical Sequence Model}

In this variation, we simply remove the reordering operations from the sequences and train the neural model only on the lexical sequences. This allows us to study how much of the improvement is obtained due 
reordering triggers integrated within these lexical sequences versus addressing sparsity by learning generalized representations. However note that such a lexical sequence model can still be considered a reordering model because the source was pre-ordered (or linearized) based on target (See Table \ref{tab:operations}) and generated in the target order. This model is based on the tuple sequence model \cite{Marino2006} and several neural variants of it are presented in \cite{le-allauzen-yvon:2012:NAACL-HLT}. 
Another variation is presented in \cite{Devlin_2014_acl}, but rather than pre-ordering the source, they select $m$ neighboring word on the left and right sides of the source word $s_i$ that is aligned to the target word $t_i$ being modeled.

\subsubsection{Decoding} 

We integrate these models as a feature in phrase-based decoding. Word alignments for the current phrase along with the history of previously generated operations are used to generate a new sequence of (lexical and reordering) operations. This sequence is then scored to give probability of the hypothesized phrase. 

\section{Experiments}
\label{sec:exp}

\subsection{Training Data} 

We experimented with German$\leftrightarrow$English language pairs using the data made available for the  International Workshop on Spoken Language Translation (IWSLT'14). The data contains roughly 5M bilingual sentence pairs. We used only TED corpus \cite{cettolo2014report} plus a subset of 800K parallel sentences from the rest of the parallel data to train the neural models.\footnote{Training models on the entire data required roughly 18 days of wall-clock time (18 hours/epoch on a Linux Ubuntu 12.04.5 LTS running on a 16 Core Intel Xeon E5-2650 2.00Ghz and 64Gb RAM) on our machines. We ran one baseline experiment with all the data and did not find it better than the system trained on randomly selected subset  of the data.
In the interest of time, we therefore reduced the NN training to a subset (1M).} We concatenated dev- and test-2010 for tuning and used test2011-2013 for evaluation. 

\subsection{MT Settings} 

We trained a Moses phrase-based system \cite{Moses:2007} following the settings described in \cite{birch-etal:2014:IWSLT}:  maximum sentence length of 80, Fast-align \cite{dyer-chahuneau-smith:2013:NAACL-HLT} for word-alignments, an interpolated Kneser-Ney smoothed 5-gram language model \cite{Heafield-kenlm}, lexicalized reordering \cite{koehn2005epc} and a 5-gram OSM model \cite{durrani-EtAl:2013:Short}. We used k-best batch MIRA \cite{cherry-foster:2012:NAACL-HLT} for tuning.\footnote{All systems were tuned twice.} We trained alternative baselines by adding OSM models trained on POS and word clusters (50) obtained by running {\tt mkcls} \cite{durrani-EtAl:2014:Coling}. We used LoPar 
for German and MXPOST tagger 
for English POS tags. We trained 7-gram models to enable wider context than the regular word-based models.

\subsection{NN Training} 

We trained our neural reordering models using NPLM\footnote{http://nlg.isi.edu/software/nplm/} toolkit \cite{Vaswani_2013_emnlp} with the following settings. We used a target context of 6 words (including reordering operations) and a corresponding source window of 7 words (also including reordering operations), forming a joint stream of 14-grams for training. We restricted source and target side vocabularies to 20K and 40K most frequent words. We used an input embedding layer of 150 and an output embedding layer of 750. Only one hidden layer is used with a Noise Contrastive Estimation\footnote{Training neural language model with backpropagation could be prohibitively slow because for each training instance, the softmax layer requires a summation over the entire output vocabulary. One way to avoid this repetitive computation is to use a Noise Contrastive Estimation of the loss function.} or NCE \cite{Gutmann_2010_jmlr}. Training was done using mini-batch size of 1000 and using 100 noise samples. All models were trained for 25 epochs. 

\begin{table}

\centering
\begin{tabular}{l|lll|l}
\toprule
 \multicolumn{5}{c}{\bf German-English} \\
 \midrule
System & test11 & test12 & test13 & Avg.\\
\midrule
Baseline & 35.0 &  30.3   &  27.1 & 30.8 \\
OSM$_{pos}$ & 35.3 & 30.5  & 27.1 & 31.0 \\
OSM$_{mkcls}$ & 35.1 & 30.1 & 26.8  & 30.7 \\
\midrule
OSM$_{neural}$ & 35.8 & 31.5 & 27.0 & 31.4 \\
Lex.reo$_{neural}$ & 35.5 & 31.1  & 27.2 & 31.3 \\
Lex$_{neural}$ & 35.3 & 30.8  & 26.9 & 31.0  \\
\midrule
 \multicolumn{5}{c}{\bf English-German} \\
 \midrule
Baseline & 25.7 &  21.7  &  23.4 &  23.6 \\
OSM$_{pos}$   & 25.9 & 21.9 & 23.8 & 23.9 \\
OSM$_{mkcls}$   & 25.8 &  21.8 & 23.4 & 23.7 \\
\midrule
OSM$_{neural}$ & 26.1 & 22.1 & 24.2 & 24.1 \\
Lex.reo$_{neural}$ & 26.1 & 22.4  & 23.7 &  24.1 \\
Lex$_{neural}$ & 26.0 &  22.2 & 23.7 & 24.0  \\
\bottomrule
\end{tabular}
\caption{\label{tab:nosm} Comparing performance of Neural Reordering Models against N-gram-based Models. Quality measured in cased-bleu \cite{Papineni:Roukos:Ward:Zhu:2002} }
\end{table}

\subsection{Results}

Table \ref{tab:nosm} compares the results for our neural reordering models against baseline containing traditional reordering models. The baseline system is equipped with lexicalized and OSM model trained over word forms using count-based/n-gram-based models. We see that adding OSM models trained over generalized representation such as POS tags help slightly (+0.2 BLEU improvement in DE-EN and +0.3 in EN-DE). Using word clusters instead of POS tags did not help as much. 

The next set of rows show results when using neuralized OSM and Lexicalized reordering models. The neural OSM model gave an improvement of +0.6 and +0.5 in DE-EN and EN-DE pairs. Neuralized lexical reordering performed almost as good as the neural OSM model suggesting that fine-grained reordering tags and hierarchical jumps add little value. The lexical sequence model without reordering tags (last row) performed  lower (in the DE-EN pair) showing that there is some value in integrating reordering tags\footnote{We also tried variations with reordering tags either on source or target side. The current variation with tags on both sides worked best.} during generation. In the EN-DE pair the difference is insignificant showing that much of the gains are coming from addressing lexical sparsity and not better reordering.

\section{Neural Machine Translation}
\label{sec:nmt}

Neural Machine Translation \cite{bahdanau:ICLR:2015,sutskever2014sequence} is quickly becoming the predominant approach to machine translation. Rather than modeling different linguistic aspects (lexical generation, reordering, fertility etc.) as feature components and tuning them to optimize BLEU, NMT is trained in an end-to-end fashion.  Given a bilingual sentence pair, we first generate a vector representation of the source sentence using {\tt encoder} and then map this vector to target sentence using a {\tt decoder}. The long distance source and target contextual dependencies are modeled using recurrent neural networks (RNN) with bilingual Long Short Term Memory (LSTM) \cite{hochreiter1997long}. The attention model \cite{bahdanau:ICLR:2015} serves as an alignment model which enables the decoder to focus on different parts of the source as it generates the target sentence. Unlike phrase-based decoding, the reordering window is not limited to a frame of 6 words. This allows NMT to capture very long range reordering like syntax-based models \cite{galley-EtAl:2006:COLACL}. 

In this work, we tried to explore whether explicitly integrating reordering triggers into the RNN-based {\tt encoder} and {\tt decoder}, improve the performance of the attentional model. We use the training data generated earlier (to train the neural OSM models -- See Table \ref{tab:operations}), to train the {\tt sequence-to-sequence} NMT model. This allows the {\tt decoder} to condition on both lexical and reordering states when generating the new target word, which itself can be a  word or a reordering operation. Our motivation was that such reordering triggers and pre-ordering of source\footnote{Remember that we linearize the source based on target using word-alignments} might help the attention mechanism with its task.

Note that the target sequence and alignments are both latent variables during decoding, we need to predict the pre-ordered (or reordering augmented sequence). To do this, we additionally train a source$\rightarrow$pre-ordered (or reordering augmented) source sequence using another {\tt sequence-to-sequence} model.

\subsection{System Settings} 

We trained a 2-layered LSTM encoder-decoder with attention. We used {\tt seq2seq-attn} implementation \cite{kim2016} with the following settings: word vectors and LSTM states have 500 dimensions, SGD with initial learning rate of 1.0 and rate decay of 0.5, and dropout of 0.3. The MT systems are trained for 20 epochs, and the model with best dev loss is used for extracting features for the classifier. 

\begin{table}

\centering
\begin{tabular}{l|lll|l}
\toprule
 \multicolumn{5}{c}{\bf German-English} \\
 \midrule
System & test11 & test12 & test13 & Avg.\\
\midrule
Baseline & 33.9 &  29.2   & 27.5 & 30.2  \\
OSM & 32.2 & 27.6 & 25.6 &  28.5 \\
Lex.reo & 29.2 & 24.8 & 22.8 & 25.6 \\
Lex & 30.8 &  26.6 & 23.9 &  27.1 \\
\bottomrule
\end{tabular}
\caption{\label{tab:nmt} Training NMT systems with pre-ordered data, with lexical reo. operations, OSM operations}
\end{table}

\subsection{Results} 

Table \ref{tab:nmt} shows the results from training NMT systems from pre-ordered data and using reordering augmented data. No gains were observed compared to the baseline system. In fact there was significant drop in all cases. One reason for this drop could be inaccuracy in predicting pre-ordered (reordering augmented) sequences. This can be seen in the BLEU scores shown in  Table \ref{tab:pnmt}.\footnote{The BLEU scores are computed using pre-ordered (or reordering augmented) references generated using word-alignments of original source-target evaluation sets.} \cite{AndyWay} also found pre-ordering the source-side in Neural MT deteriorated system performance in Japanese$\leftrightarrow$English and Chinese$\leftrightarrow$English pairs. They conjectured that pre-ordering introduces noise in terms of word-order hindering the learning process more difficult.

\begin{table}

\centering
\begin{tabular}{l|lll|l}
\toprule
 \multicolumn{5}{c}{\bf German-English} \\
 \midrule
System & test11 & test12 & test13 & Avg.\\
\midrule
OSM & 45.7 & 42.0 & 36.6 &  41.4 \\
Lex.reo & 48.0 & 45.2 & 43.2 &  45.5 \\
Lex & 52.0 &  50.8 &  49.3 &  50.7 \\
\bottomrule
\end{tabular}
\caption{\label{tab:pnmt} Source to pre-ordered (or reordering augmented) system}
\end{table}

\section{Related Work}
\label{sec:rw}

A significant amount of research has been carried to alleviate data sparsity when translating into or from morphologically rich languages.  \cite{koehn-hoang:2007:EMNLP-CoNLL2007} integrated different levels of linguistic information as factors into the phrase-based translation model. The idea of translating to stems and then inflecting the stems in a separate step has been studied by several researchers \cite{toutanova-suzuki-ruopp:2008:ACLMain,fraser-EtAl:2012:EACL2012}. POS tags are used in bilingual sequence models to enable wider context by \cite{niehues-EtAl:2011:WMT,crego-yvon:2010:POSTERS,durrani-EtAl:2014:Coling}. Several researchers used word clusters in training data to obtain smoother distributions and better generalizations \cite{wuebker-EtAl:2013:EMNLP,chahuneau13morphogen,bisazza-monz:2014:Coling}. \cite{feng-cohn-du:2014:W14-16} used factors and parallel back-offs to address the issue of data sparsity. Continuous space models are used earlier for n-best re-ranking or directly as a feature in phrase-based MT \cite{schwenk:2012,le-allauzen-yvon:2012:NAACL-HLT, kalchbrenner13,gao-EtAl:2014:P14-1,Devlin_2014_acl,guta-EtAl:2015:EMNLP}. \cite{cui-wang-li:2016:N16-1} recently proposed an LSTM recurrent neural reordering model which directly models word pairs and their alignment. However, because SMT decoder requires fixed history, it is only possible to use the feature in the n-best re-ranking.  

A whole new paradigm based on deep neural network evolved as a parallel framework for machine translation \cite{bahdanau:ICLR:2015,sutskever2014sequence}. The RNN-based sequence-to-sequence model learns generalized representations to overcome data sparsity problems and learn long distance dependencies successfully. This is further enhanced by using sub-word \cite{sennrich2015neural} or character-based models \cite{kim2015character} to address the OOV-word problem. \cite{cohn-EtAl:2016:N16-1} has recently shown that integrating structural biases based on relative positions and fertilities improves the attention mechanism. \cite{sennrich-haddow-birch:2016:N16-1} and \cite{KobusCS16} used side-constraints i.e. adding suffix tag at the end of the source sentence or prefix tag  in the beginning of the target sentence to control the behavior of the decoder i.e. politeness in the case of former and domain in the latter. Our work is similar in a sense that we are trying to add reordering constraints, forcing the decoder to produce a specific reordering pattern. However, our method did not yield any improvements.

\section{Conclusion}
\label{sec:conc}

Traditional reordering models in phrase-based system suffer from data sparsity. In this paper, we presented neuralized versions of these reordering models (the OSM and Lexicalized reordering models) and used them as a feature in Phrase-based SMT. Our evaluation on German-English language pairs showed an improvement of up to 0.6 BLEU points. We also demonstrated gains compared to the previous solution where these models are trained on parts-of-speech tags and word clusters, to address data sparsity and for better generalization. The code will be pushed to Moses toolkit.\footnote{http://www.statmt.org/moses/} We also tried our pre-ordered and reordering augmented training data to train sequence-to-sequence neural MT models, with a motivation to explicitly add reordering triggers in the encoder representation and aid the attention mechanism. However, our modification to the natural source order and integration of reordering symbols in the training data, did not yield improvement. 

\section{Acknowledgements}

We would like to thank the anonymous reviewers for their useful feedback.

\bibliographystyle{IEEEtran}
\bibliography{iwslt15}
\end{document}

%% file: figure2.tex
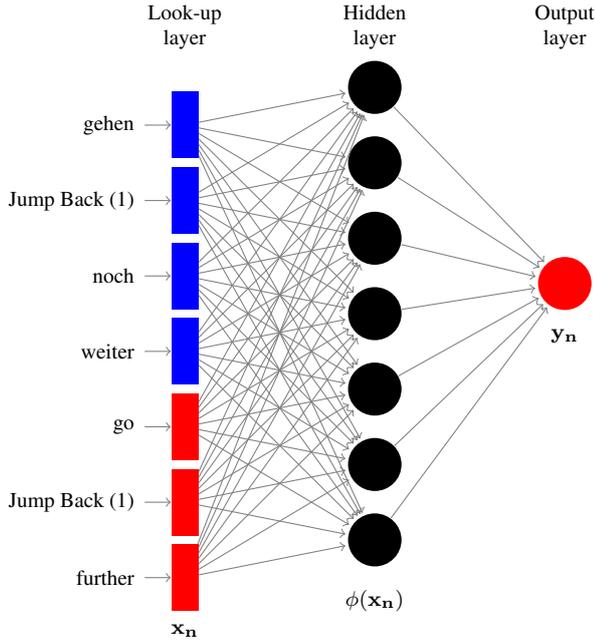
\begin{figure}
\centering
\footnotesize
\begin{tikzpicture}[shorten >=1pt,->,draw=black!50, node distance=\layersep]
    \tikzstyle{every pin edge}=[<-]

	\tikzstyle{neuron}=[circle,fill=black!50,minimum size=20pt,inner sep=0pt]
	\tikzstyle{emb1}=[rectangle,fill=blue!100,minimum width=10pt, minimum height=25pt,inner sep=0pt]
	\tikzstyle{emb2}=[rectangle,fill=red!100,minimum width=10pt, minimum height=25pt,inner sep=0pt]
	\tikzstyle{plate}=[rectangle,draw=black!100,minimum width=50pt,minimum height=55pt,thin,inner sep=0pt]

	\tikzstyle{input neuron}=[neuron, fill=green!50];
    \tikzstyle{output neuron}=[neuron, fill=red!100];
    \tikzstyle{hidden neuron}=[neuron, fill=black];
    \tikzstyle{noise node}=[neuron, fill=gray!100];
    \tikzstyle{class node}=[neuron, fill=brown!100];

	\tikzstyle{annot} = [text width=4em, text centered]


	\node[emb1, pin=left:gehen] (I-1) at (0,-1) {};
	\node[emb1, pin=left:Jump Back (1)] (I-2) at (0,-2) {};
\node[emb1, pin=left:noch] (I-3) at (0,-3) {};

	\node[emb1, pin=left:weiter] (I-4) at (0,-4) {};
	\node[emb2, pin=left:go] (I-5) at (0,-5) {};
    
    \node[emb2, pin=left:Jump Back (1)] (I-6) at (0,-6) {};
    
    \node[emb2, pin=left:further] (I-7) at (0,-7) {};

    \foreach \name / \y in {1,...,7}
        \path[yshift=0.5cm]
            node[hidden neuron] (H-\name) at (\layersep,-\y) {};

    \node[output neuron] (O) at (5, -3.1){};



    \foreach \source in {1,...,7}
        \foreach \dest in {1,...,7}
            \path (I-\source) edge (H-\dest);

    \foreach \source in {1,...,7}
        \path (H-\source) edge (O);


    \node[annot,above of=H-1, node distance=0.8cm] (hl) {Hidden layer};
    \node[annot,below of=H-7, node distance=0.8cm] {$\phi({\mathbf{x_n}})$};
    \node[annot,left of=hl] {Look-up layer};
    \node[annot,below of=I-7, node distance=0.7cm] {$\mathbf{x_n}$};
    \node[annot,right of=hl] {Output layer};
     \node[annot,below of=O, node distance=0.7cm]  {$\mathbf{y_n}$};
\end{tikzpicture}
\caption{Neural OSM model where we use $3$-gram target words and a source context window of size $4$. For illustration, the output $y_n$ is shown as a single categorical variable (scalar) as opposed to the traditional one-hot vector representation.}
\vspace{-2mm}
\label{fig:NN}
\end{figure}